\documentclass{article}

\usepackage{microtype}
\usepackage{graphicx}
\usepackage{subfigure}
\usepackage{booktabs} 
\usepackage{tabularx}
\usepackage{listings}
\usepackage{pgfplots}
\usepackage{tikz}
\usepackage{todonotes}
\usepackage{tikz-cd}
\usepackage[binary-units=true]{siunitx}
\usepackage{amsmath}
\usepackage{amssymb}

\usepackage{hyperref}

\usepackage[accepted]{icml2019}
\usetikzlibrary{arrows,chains,matrix,positioning,scopes, arrows.meta}
\tikzset{
  shift left/.style ={commutative diagrams/shift left={#1}},
  shift right/.style={commutative diagrams/shift right={#1}}
}
\tikzset{%
  >={Latex[width=2mm,length=2mm]},
            base/.style = {rectangle, rounded corners, draw=black,
                           minimum width=4cm, minimum height=1cm,
                           text centered, font=\sffamily},
  activityStarts/.style = {base, fill=blue!50},
       startstop/.style = {base, fill=red!50},
    activityRuns/.style = {base, fill=green!50},
         process/.style = {base, minimum width=2.5cm, fill=orange!55,
                           font=\ttfamily},
}
\makeatletter
\tikzset{join/.code=\tikzset{after node path={%
\ifx\tikzchainprevious\pgfutil@empty\else(\tikzchainprevious)%
edge[every join]#1(\tikzchaincurrent)\fi}}}
\makeatother
\tikzset{>=stealth',every on chain/.append style={join},
         every join/.style={->}}
\tikzstyle{labeled}=[execute at begin node=$\scriptstyle,
   execute at end node=$]
\icmltitlerunning{Stratospheric Aerosol Injection as a Deep Reinforcement Learning Problem}

\pgfplotsset{width=7cm,compat=1.8}

\begin{document}

\twocolumn[
\icmltitle{Stratospheric Aerosol Injection as a Deep Reinforcement Learning Problem}



\icmlsetsymbol{equal}{*}

\begin{icmlauthorlist}
\icmlauthor{Christian Schroeder de Witt}{equal,ox_eng}
\icmlauthor{Thomas Hornigold}{equal,ox_atmos}
\end{icmlauthorlist}

\icmlaffiliation{ox_eng}{Department of Engineering, University of Oxford, United Kingdom}
\icmlaffiliation{ox_atmos}{Department of Atmospheric, Oceanic and Planetary Physics, University of Oxford, United Kingdom}

\icmlcorrespondingauthor{Christian Schroeder de Witt}{schroederdewitt@gmail.com}
\icmlcorrespondingauthor{Thomas Hornigold}{thomasa2z@hotmail.com}

\icmlkeywords{Machine Learning, ICML, Climate Science, Deep Reinforcement Learning, Geoengineering, Climate Modeling, Stratospheric Aerosol Injection}

\vskip 0.3in
]

\printAffiliationsAndNotice{\icmlEqualContribution} 

\begin{abstract}
As global greenhouse gas emissions continue to rise, the use of stratospheric aerosol injection (SAI), a form of solar geoengineering, is increasingly considered  in order to artificially mitigate climate change effects. However, initial research in simulation suggests that naive SAI can have catastrophic regional consequences, which may induce serious geostrategic conflicts. Current geoengineering research treats SAI control in low-dimensional approximation only. We suggest treating SAI as a high-dimensional control problem, with policies trained according to a context-sensitive reward function within the Deep Reinforcement Learning (DRL) paradigm.  In order to facilitate training in simulation, we suggest to emulate HadCM3, a widely used General Circulation Model, using deep learning techniques. We believe this is the first application of DRL to the climate sciences.
\end{abstract}

\section{Introduction}
\label{sec:introduction}

As global greenhouse gas emissions continue to rise, the use of geoengineering in order to artificially mitigate climate change effects is increasingly considered. Stratospheric aerosol injection (SAI), which reflects incoming solar radiative forcing and thus can be used to offset excess radiative forcing due to the greenhouse effect, is widely regarded as one of the most technically and economically feasible methods \cite{Crutzen2006,macmartin_geoengineering:_2014,smith_stratospheric_2018}. However, naive deployment of SAI has been shown in simulation to produce highly adversarial regional climatic effects in regions such as India and West Africa  
 \cite{ricke_regional_2010}. Wealthy countries would most likely be able to trigger SAI unilaterally, i.e. China, Russia or the US could decide to fix their own climates and disrupt the ITCZ, which influences the monsoon over India, as collateral damage. If geoengineering is ceased before the anthropogenic radiative forcing it sought to compensate for has declined, termination effects with rapid warming would result \cite{doi:10.1002/jgrd.50762}. Understanding both how SAI can be optimised and how to best react to rogue injections is therefore of crucial geostrategic interest \cite{YU201510}. 
 
In this paper, we argue that optimal SAI control can be characterised as a high-dimensional Markov Decision Process (MDP) \cite{bellman_dynamic_1957}. This motivates the use of deep reinforcement learning (DRL) \cite{mnih_human-level_2015} in order to automatically discover non-trivial, and potentially time-varying, optimal injection policies or identify catastrophic ones.
 To overcome the inherent sample inefficiency of DRL, we propose to emulate a Global Circulation Model (GCM) using deep learning techniques. To our knowledge, this is the first proposed application of deep reinforcement learning to the climate sciences.

\section{Related work}
\label{sec:related}

General Circulation Models (GCMs), which simulate the earth's climate on a global scale, are inherently computationally intensive. Simple statistical methods are routinely used in order to estimate climate responses to slow forcings \cite{castruccio_statistical_2013}. Recently, the advent of deep learning has led to a number of successful emulation attempts of full GCMs used for weather prediction \cite{duben_challenges_2018}, as well as for sub-grid scale processes \cite{brenowitz_prognostic_2018,rasp_deep_2018}, including precipitation \cite{ogorman_using_2018}. This suggests that the emulation of the response of regional variables, such as precipitation and surface temperature, to aerosol injection forcings may now be within reach.

Investigation of optimal SAI control within the climate community is currently constrained to low-dimensional injection pattern parametrisations \cite{ban2010geoengineering} or manual grid search over edge cases of interest \cite{Jackson2015}. Even in simple settings, it has been shown that regional climate response is sensitive to the choice of SAI policy \cite{Macmartin2013}. In addition, super-regional impacts on El Nino/Southern Oscillation have been demonstrated \cite{acp-15-11949-2015}. This suggests that climate response to SAI is sensitive enough to warrant a high-dimensional treatment.

Altering the injection altitude, latitude, season, or particle type - possibly even with the use of specially engineered photophoretic nanoparticles \citep{keith_photophoretic_2010} - may provide the ability to "tailor" fine-grained SAI. But, presently, stratospheric aerosol models have substantially different responses to identical injection strategies \cite{pitari_stratospheric_2014}, suggesting directly simulating the implications of these strategies - and the range of aerosol distributions that can be attained - requires further model development.

\begin{figure*}[h!]
\centering
\begin{minipage}[c]{0.24\textwidth}
\centering
    \includegraphics[width=\textwidth, height=1.5in]{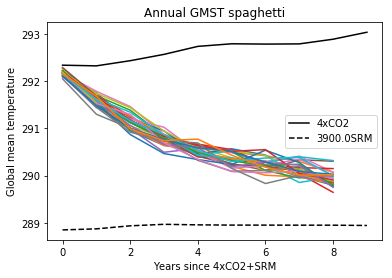}
\end{minipage}
\begin{minipage}[c]{0.24\textwidth}
\centering
    \includegraphics[width=\textwidth, height=1.5in]{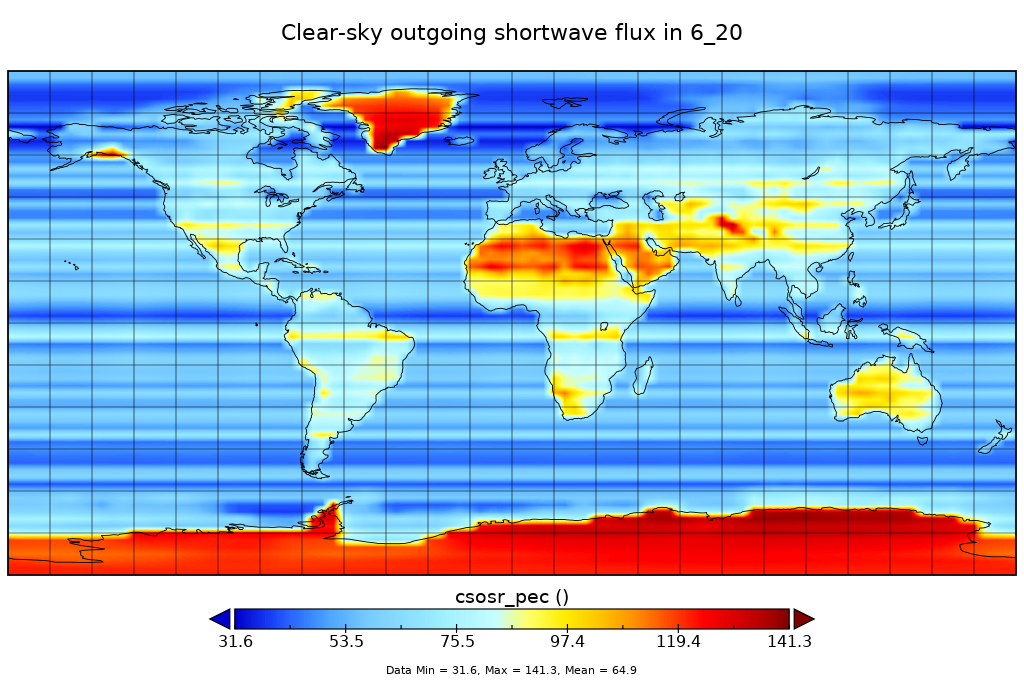}
\end{minipage}
\begin{minipage}[c]{0.24\textwidth}
\centering
    \includegraphics[width=\textwidth, height=1.5in]{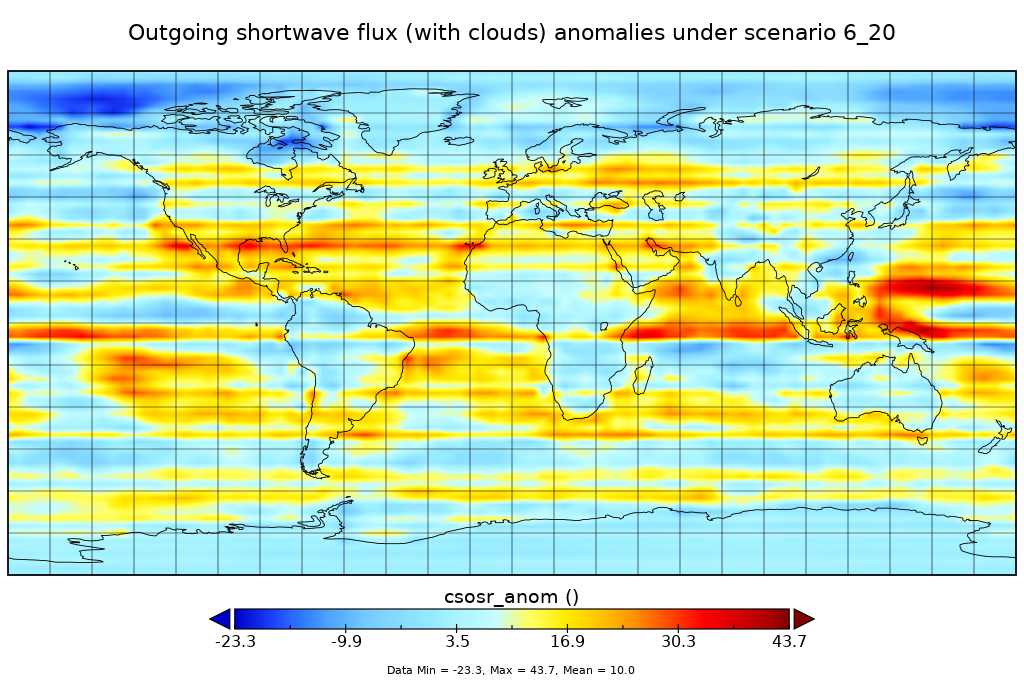}
\end{minipage}
\begin{minipage}{0.24\textwidth}
\centering
    \includegraphics[width=\textwidth, height=1.5in]{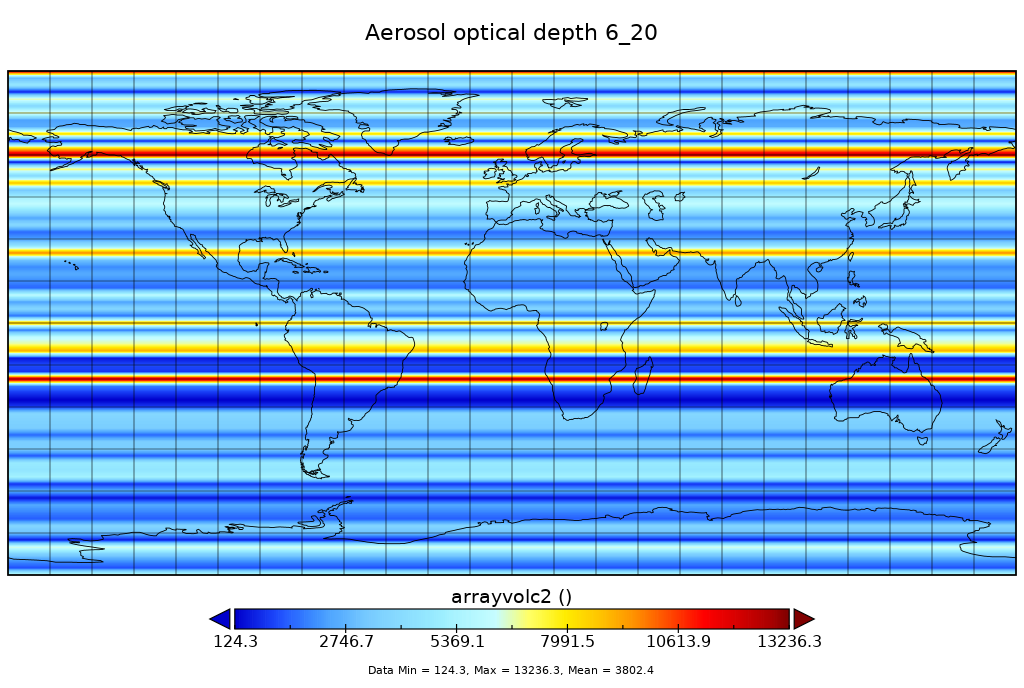}
\end{minipage}
\caption{from left: 1. Global mean cooling relative to $\rho$-uniform and zero AOD baselines (full HadCM3 dataset) 2. Clear-sky upwelling shortwave radiative flux 3. Outgoing shortwave radiative flux (with clouds) 4. AOD distribution (2., 3. and 4. at same time / same run) }
\label{fig:sky}
\end{figure*}

\section{GCM emulation}
\label{sec:GCM}
We use HadCM3 \cite{gordon_simulation_2000} to simulate the climate response to SAI as it is the first GCM not to require flux adjustments to avoid large scale climate drift. In addition, HadCM3 is still used as a baseline model for IPCC reports \cite{ipcc_fifth_2013}.

The radiative forcing of sulfate aerosols is emulated in HadCM3 by adjusting the aerosol optical depth (AOD) in the lower stratosphere, i.e. a larger AOD corresponds to a larger sulfate aerosol concentration. Predominantly zonal winds in the stratosphere are assumed to keep aerosol optical depth zonally uniform to first order, so it is prescribed for each of the model's $73$ latitude bands. We also assume that aerosol concentration completely decays within a year and that aerosol concentration is upper-bounded by coagulation effects \cite{hansen_efficacy_2005} and thus capped at $4\overline{\rho}$ in each latitude band, with $\overline{\rho}=m_oA_S^{-1}$, where $A_S$ is the surface area of the lower stratosphere band at an altitude of $20\ km$ \cite{smith_stratospheric_2018}.

Despite being up to a factor $10^3$ faster than many contemporary GCMs, a single HadCM3 year still corresponds to about $15$ hours of computation on a generic single-thread CPU. In order to employ deep reinforcement learning, we therefore require a fast emulator that can predict next states in a matter of milliseconds.

We approximate the full HadCM3 state $s_t$ at time $t$ by the scalar surface fields \textit{sea ice fraction} $S_t(x,y)$, \textit{surface temperature} $T_t(x,y)$, \textit{depth layer-weighted ocean heat content} $H_t(x,y)$ and \textit{stratospheric aerosol optical depth} $\tau_t(x,y)$. From these quantities, the emulator needs to predict $S_{t+1},H_{t+1}$ and $T_{t+1}$, as well as other quantities of interest to the policy optimisation objective, such as local precipitation rates $P_{t+1}(x,y)$.  All these quantities are returned from HadCM3 simulations as scalar grids of dimension $73\times 96$.

To emulate HadCM3, we use an encoder-decoder network similar to UNet \cite{ronneberger_u-net:_2015} given HadCM3 output is largely deterministic, which can be augmented with an prediction uncertainty channel. 
We pre-train the encoder on ImageNet \cite{mirowski_learning_2016} and fine-tune the output layers on  $2000$ output samples of HadCM3 rollouts based on aerosol density distributions drawn randomly from a $73$-dimensional Dirichlet distribution with shape parameters $\alpha_k=1.5$ (to discourage extremes) and output scaling factor $\overline{\rho}$. We reject samples violating the $4\overline{\rho}$ coagulation cap.

Preliminary simulation results suggest that emulator training would likely benefit from auxiliary tasks \cite{liebel_auxiliary_2018} related to cloud cover prediction (see Figure \ref{fig:sky}).

\section{Reinforcement learning setting}
\label{sec:RL}

GCM emulator states $s_t$ and sequential aerosol injections conditioned thereon together form a Markov Decision Process \cite{bellman_dynamic_1957}. At each time step, which amounts to a year of simulated climate, the agent decides how much aerosol to inject into each of $73$ evenly spaced latitude bands, overall selecting an action $\mathbf{u}_t\in\mathbb{R}^{73}_+$. The environment then returns a scalar reward $r_t$ as feedback to the agent. Optimal injection policies $\pi(\mathbf{u}_t|\mathbf{s}_t)$ are then learnt by maximizing the expected future-discounted cumulative reward $R_t=\sum^T_{t=0}\gamma^tr_t$, where $\gamma\in[0,1]$ is a discount factor and $T=10$ corresponds to an episode length of $10$ years.
  
For reasons of simplicity and robustness, we employ an off-policy deep Q-learning \cite{mnih_human-level_2015} approach and discretise the action space using $n_u=10$ bins of equal size for each latitude band. As the resulting joint action space is large ($10^{73}$), we factorise the joint state-action value function. This could either be achieved using low-dimensional embeddings \cite{dulac-arnold_deep_2015} or, more robustly, by factorising the joint value function using techniques originally developed for cooperative multi-agent settings \cite{sunehag_value-decomposition_2018,rashid_qmix:_2018}. Value function network architecture is based on a convolutional encoder similar to the one used by the GCM emulator.

A simple choice for an upper-bounded reward function $r_t$ that discourages extreme changes in regional climate is
$$- \max_{x,y\in A} \left[\alpha_P|\Delta^t_P(x,y)| + \alpha_T|\Delta^t_T(x,y)|\right]$$
where $\Delta^t_P$ is the difference between the regional precipitation rate and its pre-industrial average (similarly $\Delta^t_T$ for surface temperature), $A$ is the earth's surface grid and $\alpha_P,\alpha_T>0$ are scalar hyperparameters.
More advanced reward functions might be weighted by additional local factors, including population density, local resilience factors and other aspect of eco-socio-economic interest and skewed according to measures such as climate adaptability and regional climate specifics.

To ensure physical consistency and robustness, SAI control policies learnt within the emulator are subsequently cross-verified in HadCM3 and/or other GCMs.

\section{Conclusion and Outlook}
We propose the study of optimal SAI control as a high-dimensional control problem using a fast GCM emulator and deep reinforcement learning.

We believe that DRL may become an important tool in the study of SAI and other geoengineering approaches, such as marine cloud brightening, over the next decade.

\bibliographystyle{icml2019}
\bibliography{final}
\end{document}